\documentclass{article}

\pdfoutput=1

\usepackage{amsmath,amsfonts,bm}

\def\eqref#1{equation~\ref{#1}}

\def\1{\bm{1}}

\DeclareMathAlphabet{\mathsfit}{\encodingdefault}{\sfdefault}{m}{sl}
\SetMathAlphabet{\mathsfit}{bold}{\encodingdefault}{\sfdefault}{bx}{n}

\usepackage[final]{neurips_2020}

\setcitestyle{numbers,square,super}
\usepackage[utf8]{inputenc} 
\usepackage[T1]{fontenc}    
\usepackage{hyperref}       
\usepackage{url}            
\usepackage{booktabs}       
\usepackage{amsfonts}       
\usepackage{nicefrac}       
\usepackage{microtype}      
\usepackage{xcolor}         
\usepackage{xspace}
\usepackage{amsmath}
\usepackage{graphicx}
\usepackage{dsfont}
\usepackage{amssymb}
\usepackage{algorithm2e}
\usepackage{multirow}
\usepackage{caption}
\usepackage{wrapfig}
\graphicspath{
{image/}
}

\newcommand{\xl}[1]{{\color{orange}[xl: #1]}}
\newcommand{\yujie}[1]{{\color{purple}#1}}
\newcommand{\note}[1]{{\color{green}[yj: #1]}}

\newcommand{\revise}[1]{{\color{black}#1}}

\newcommand{\name}{SinGRAV\xspace}

\title{\name: Learning a Generative Radiance Volume from a Single Natural Scene}

\author{
\centering
\setlength{\tabcolsep}{37pt}
\begin{tabular}{cc}
    Yujie Wang$^{1,3}$ & Xuelin Chen$^{2}$  \\
    $^1$Shandong University &$^2$Tencent AI Lab  \\
    \texttt{yujiew.cn@gmail.com} & \texttt{xuelin.chen.3d@gmail.com}  \\
    \multicolumn{2}{c}{} \\
    \multicolumn{2}{c}{Baoquan Chen$^3$} \\
    \multicolumn{2}{c}{$^3$Peking University}\\
    \multicolumn{2}{c}{\texttt{baoquan.chen@gmail.com}}\\
    \end{tabular}%
}

\begin{document}

\maketitle


\begin{abstract}
We present a 3D generative model for general natural scenes.
Lacking necessary volumes of 3D data 
characterizing the target scene,
we propose to learn from 
a single scene.
Our key insight is that
a natural scene often contains multiple constituents whose geometry, texture, and spatial arrangements follow some clear patterns, 
but still exhibit rich variations over different regions within the same scene.
This suggests localizing the learning of a generative model on substantial local regions.
Hence, we exploit a multi-scale convolutional network, which possesses the spatial locality bias in nature, to learn from the statistics of local regions at multiple scales within a single scene.
In contrast to existing methods,
our learning setup bypasses the need to collect data from many homogeneous 3D scenes for learning common features.
%
We coin our method \name, for learning a \emph{G}enerative \emph{RA}diance Volume from a \emph{Sin}gle natural scene.
We demonstrate the ability of \name in generating plausible and diverse variations from a single scene,
the merits of \name over state-of-the-art generative neural scene methods,
as well as the versatility of \name by its use in a variety of applications,
spanning 3D scene editing, composition, and animation.
Code and data will be released to facilitate further research.
\end{abstract}


\section{Introduction}

%
Recently,
3D generative modeling has made great strides via gravitating towards neural scene representations,
which boast unprecedentedly photo-realism.
Generative neural scene models~\citep{graf, giraffe, eg3d, stylenerf, pi-gan} can now draw class-specific realistic scenes (e.g., cars and portraits),
offering a glimpse into the boundless universe in the virtual.
%
Yet an obvious question
is how we can go beyond class-specific scenes, and achieve similar success with \emph{general} \emph{natural} scenes, creating \emph{at scale} diverse scenes of more sorts.
This work presents, to our knowledge, the first endeavor towards answering this question.

\begin{figure*}[t]
\begin{center}

\begin{tabular}{ccc}
\vspace{-0.4mm}
\makebox[0.25\textwidth][c]{\small{Training scene}} & \multicolumn{2}{c}{\small{Random generation}} \\
\multicolumn{3}{c}{\includegraphics[width=0.985\linewidth]{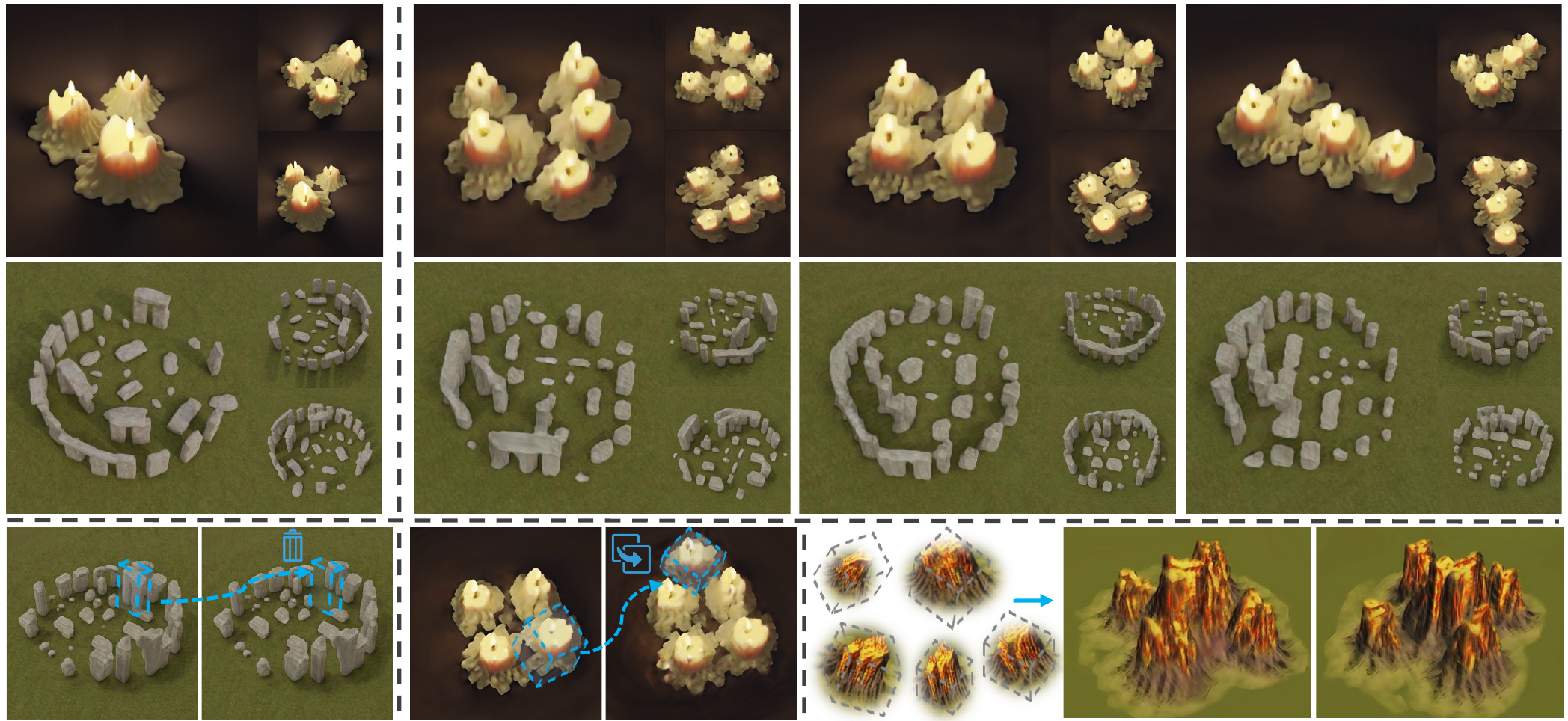}}\\
\makebox[0.25\textwidth][c]{\small{Editing - removal}} & \makebox[0.25\textwidth][c]{\small{Editing - duplicate}} & \makebox[0.45\textwidth][c]{\small{Composition}} \\
\vspace{-7mm}
\end{tabular}
\end{center}
\caption{
Top two rows:
From observations of a single natural scene, 
we learn a generative model to synthesize highly plausible variations.
Three views rendered from each of input/generated scenes are presented.
Note how the global and object configurations vary in generated samples, yet still resemble the original. 
Bottom: Applications enabled by \name, including removal (left) and duplicate (middle) operation for editing a 3D scene sample, and scene composition (right) that combines 5 different generated samples to form a novel complex scene.
    } 
 \label{fig:teaser}
\end{figure*}

While neural scene generation has boosted the field via learning from images with differentiable projections,
there still exists strong dependence on datasets containing images of many homogeneous scenes.
Collecting homogeneous data for each scene type ad hoc is cumbersome, and would become prohibitive when the scene type of interest varies dynamically.
%
Fortunately, parallels can be drawn following nature,
who implicitly maintains a rich variety of natural scenes.
In nature, no scenes ever stayed absolutely unaltered ---
imagine, as the stars shift, a river changes its current and little pebbles get scattered and weathered variously.
Analogically, in scene creation,
%
new scenes can be created via "varying" an existing one, yet still resembling the original (see Figure~\ref{fig:teaser}).

\setlength{\tabcolsep}{1pt}
\renewcommand{\arraystretch}{1}

Following this spirit,
we present \name,
for a \emph{generative radiance volume}
learned to synthesize variations
from
a \emph{single} general natural scene.
Building upon recent advances in 3D generative modeling,
\name also learns from visual observations of the target scene via differentiable volume rendering,
but the training supervision, scene representation, and network architecture all have to be altered fundamentally to meet unique challenges arising from learning with a single scene. 
Predominantly, 
the supervision comes from merely a single scene,
which would manifestly elude state-of-the-art models that learn common features across homogeneous scenes.

%

Natural scenes often contain multiple constituents whose geometry, appearance, and spatial arrangements follow some clear patterns, 
but still exhibit rich variations over different regions within the same scene
.
This naturally inspires one to learn from the statistics of internal regions via \emph{localizing} the training.
%
%
As our supervision comes from 2D images,
learning internal distributions consequently
grounds \name on the assumption that multi-view observations share a consistent internal distribution for learning,
which can be simply realized by uniformly placing the training cameras around the scene to obtain homogeneous images.
%
On the other hand,
%
MLP-based representations tend to synthesize holistically and perform better at modeling global patterns over local ones~\citep{repmlp},
thus favor modeling class-specific distributions.
Hence, we resort to convolutional operations,
which generate discrete radiance volumes from noise volumes with limited receptive fields,
for learning local properties over a confined spatial extent,
granting better out-of-distribution generation in terms of the global configuration.
%
Moreover, 
we adopt a multi-scale architecture containing a pyramid of convolutional GANs
to capture the internal distribution at various scales, alleviating the notorious mode-collapse issue.
%
%
This framework is similar in spirit with~\cite{singan, ingan},
however, 
important designs must be incorporated to \emph{efficiently} and \emph{effectively}
improve the plausibility of the spatial arrangement and the appearance realism of the generated 3D scene.
We demonstrate \name enables us to easily generate plausible variations of the input scene in large quantities and varieties, with novel geometry, textures and configurations.
Plausibility evaluations are conducted via perceptual studies,
and comparisons are made to state-of-the-art generative models.
The importance of various network design choices
are validated.
Finally, we show the versatility of \name by its use in a series of applications,
spanning 3D scene editing, retargeting, and animation.
\section{Related work}
%

\paragraph{Neural scene representation and rendering.}
In recent years, 
neural scene representations
have been the de facto infrastructure in several tasks, including
representing shapes \citep{imnet, deepsdf, levelset, nglod, acorn}, 
novel view synthesis \citep{hologan, nerf, synsin}, 
and 3D generative modeling \citep{blockgan, graf, pi-gan, giraffe, stylenerf, unconstrained, eg3d}. 
%
Paired with differentiable projection functions,
the geometry and appearance of the underlying scene can be optimized based on the error derived from the downstream supervision signals.
\cite{imnet, deepsdf, disn, nglod, levelset} adopt neural implicit fields to represent 3D shapes and attain highly detailed geometries. 
On the other hand,
\cite{deepvoxel, deferrender, synsin} work on discrete grids, UV maps, and point clouds, respectively,
with attached learnable neural features that can produce pleasing novel view imagery. 
More recently, 
the Neural Radiance Field (NeRF) technique~\citep{nerf} has revolutionized several research fields with a trained MLP-based radiance and
opacity field,
achieving unprecedented success in producing photo-realistic imagery. 
An explosion of NeRF techniques occurred in
the research community since then that improves the NeRF in various aspects of the problem~\citep{neuralsparse, rebain2020derf,  nerfpp, lindell2021autoint, nex, nerfw, lin2021barf, nerfmm, mixturevolume}.
A research direction drawing increasing interest, which we discuss in the following,
is to incorporate such neural representations to learn a 3D generative model possessing photo-realistic viewing effects.
\paragraph{Generative neural scene generation.}
With great success achieved in 2D image generation tasks, 
research have taken several routes to
realize 3D-aware generative models.
%
The heart of these methods is
a 3D neural scene representation, paired with the volume rendering, makes the whole pipeline differentiable with respect to the supervision imposed in image domain. 
\cite{graf, pi-gan} integrate a neural radiance field into generative models,
and directly produce the final images via volume rendering, 
yielding a 3D field modeled with great view-consistency under varying views.
To overcome the low query efficiency and high memory cost issues, 
researchers proposed to adopt 2D neural renders to improve the inference speed, and achieved high-resolution outputs~\citep{giraffe, stylenerf, eg3d}.
\cite{giraffe} utilizes multiple radiance sub-fields to represent a scene,
and shows its potential to model scenes containing multiple objects on synthetic primitive datasets.
More often than not, 
these methods are demonstrated on single-object scenes. 
\cite{unconstrained} found that the capacity of a generative model conditioned on a global latent vector is limited, 
and instead propose to use a grid of locally conditioned radiance sub-fields to model more complicated scenes. 
All these work learn category-specific models, requiring training on sufficient volumes of images data collected from many homogeneous scenes.
In this work, we target general natural scenes,
which in general possess intricate and exclusive characteristic,
suggesting difficulties in collecting necessary volumes of training data and rendering these data-consuming learning setups intractable.
Moreover, as aforementioned,
our task necessitates localizing the training over local regions, which is lacking in MLP-based representations,
leading us to the use of voxel grids in this work.

\section{Method}

\newcommand{\volume}{ V }
\newcommand{\xyz}{ \textbf{p} }
\newcommand{\density}{ \sigma }
\newcommand{\rgb}{ \textbf{c} }
\newcommand{\ray}{ \textbf{r} }
\newcommand{\expo}{ \text{exp} }
\newcommand{\Gtwod}{ G^{2D} }
\newcommand{\Gthreed}{ G^{3D} }
\newcommand{\G}{ G  }

\newcommand{\imset}{ \mathcal{X} }
\newcommand{\Gset}{ \mathcal{G} }
\newcommand{\im}{ x }
\newcommand{\depth}{d}
\newcommand{\depgen}{\tilde{\depth}}
\newcommand{\imgen}{ \tilde{\im} }

\newcommand{\volgen}{ \tilde{V} }

\newcommand{\scale}{ n }

\newcommand{\loss}{ \mathcal{L} }

\name learns a powerful generative model for generating neural radiance volumes from multi-view observations $\imset = \{ \im^1, ..., \im^m \} $ of a single scene.
In contrast to learning class-specific priors, 
our specific emphasis is to learn the internal distribution of the input scene.
\begin{figure*}
    \includegraphics[width=\linewidth]{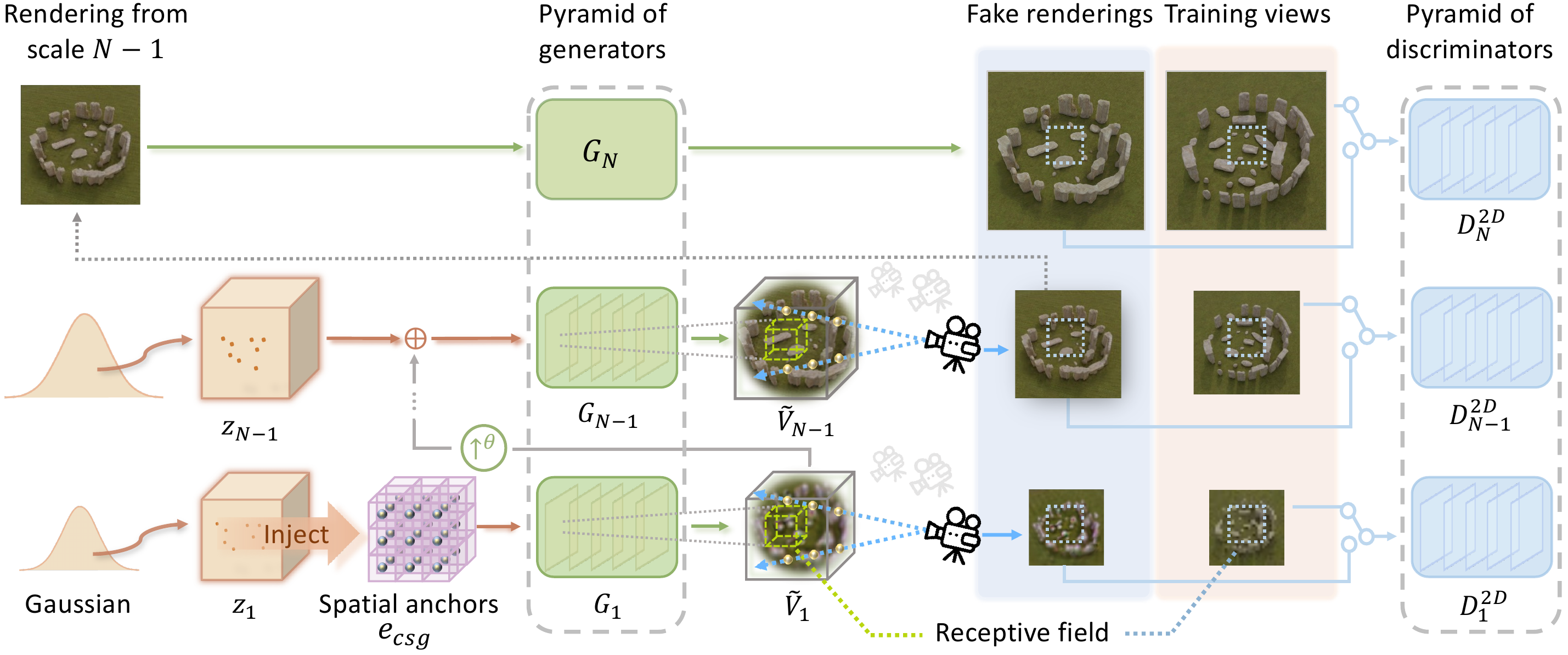}
    \caption{\name training setup. 
    A series of convolutional generators are trained to generate a scene in a coarse-to-fine manner.
    At each scale,
    $G_n$ learns to
    form a volume via
    generating realistic 3D overlapping patches, that collectively contribute to a volumetric-rendered imagery indistinguishable from the observation images of the input scene by the discriminator $D_n$.
    At the finest scale, the generator $G_N$ operates purely on 2D domain to super-resolve the imagery produced from the previous scale (top),
    significantly reducing the computation overhead.
    %
    }
    \label{fig:framework}
\end{figure*}
%
%
%
To this end,
we 
%
resort to convolutional networks, 
which inherently possess spatial locality bias,
with limited receptive fields for learning over a variety of local regions within the input scene.
%
%
The generative model is learned via adversarial generation and discrimination through 2D projections of the generated volumes.
During training, 
the camera pose is randomly selected from the training set.
We will omit the notion of the camera pose for brevity.
Moreover, we use a multi-scale framework to learn properties at different scales, ranging from global configurations to local fine texture details. 
Figure~\ref{fig:framework} presents an overview.

\subsection{Neural radiance volume and rendering}

The generated scene is represented by a discrete 3D voxel grid, and is to be produced by a 3D convolutional network.
Each voxel center stores a 4-channel vector that contains a density scalar $\density$ and a color vector $\rgb$. 
Trilinear interpolation is used to define a continuous radiance field in the volume.
We use the same differentiable function as in NeRF for volume-rendering generated volumes.
The expected color $\hat{C}$ of a camera ray $\ray$ is approximated by integrating over $M$ samples spreading along the ray:
\begin{equation}
\hat{C}{(\ray)} = \sum_{i=1}^{M} T_i \big( 1 - \expo(-\density_i \delta_i) \big) \rgb_i,
        \; \text{and} \; T_i = \expo\Big( -\sum_{j=1}^{i-1}\density_j \delta_j \Big),
\end{equation}
where
the subscript denotes the sample index between the near $t_n$ and far $t_f$ bound,
$\delta_i = t_{i+1} - t_i$ is the distance between two consecutive samples,
and $T_i$ is the accumulated transmittance at sample $i$, which is obtained via integrating over the preceding samples along the ray.
This color rendering is reduced to traditional alpha compositing with alpha values $\alpha_i = 1 - \expo(-\density_i \delta_i)$.
This function is differentiable and enables updating the volume based on the error derived from supervision signals.

\subsection{Hybrid multi-scale architecture}
We use the multi-scale framework widely followed in image generation.
%
In addition,
\name adopts a hybrid framework, which contains a series of 3D convolutional generators $\{G_{\scale}\}_{\scale=1}^{N-1}$ and a lightweight 2D convolutional generator $G_N$ (see Figure~\ref{fig:framework}). 
Specifically, 3D generators $\{G_{\scale}\}_{\scale=1}^{N-1}$ sequentially generate a volume with an increasing resolution at $N-1$ coarser scales,
with
the volume resolution increased by a factor of $\theta$ between two consecutive scales and the rendering resolution
increases by a factor of ${\mu}_r$. 
At $N$th scale, to address the overly high computation cost, 
we use a lightweight 2D generator $G_N$ to super-resolve the imagery from the preceding scale by a factor ${\mu}_s$,
achieving higher-resolution outputs. 
%
%
%
Essentially,
each generator $G_\scale$ learns to generate realistic outputs to fool an associate 2D discriminator $D^{2D}_\scale$, 
that is designed to distinguish the generated renderings from real ones.
Importantly, the generators and discriminators are designed to be equipped with \emph{limited receptive fields}, preventing them from over-fitting to the whole scene.
%

Starting from a coarsest volume,
the generation sequentially passes through all generators and increases the resolution up to the finest, 
with noise injected at each scale. 
At the coarsest scale, 
the volume generation is purely generative, i.e.\ $G_1$ maps a Gaussian noise volume $z_1$ to a radiance volume. 
\begin{wrapfigure}{r}{3.8cm}
  \begin{center}
    \includegraphics[width=3.5cm]{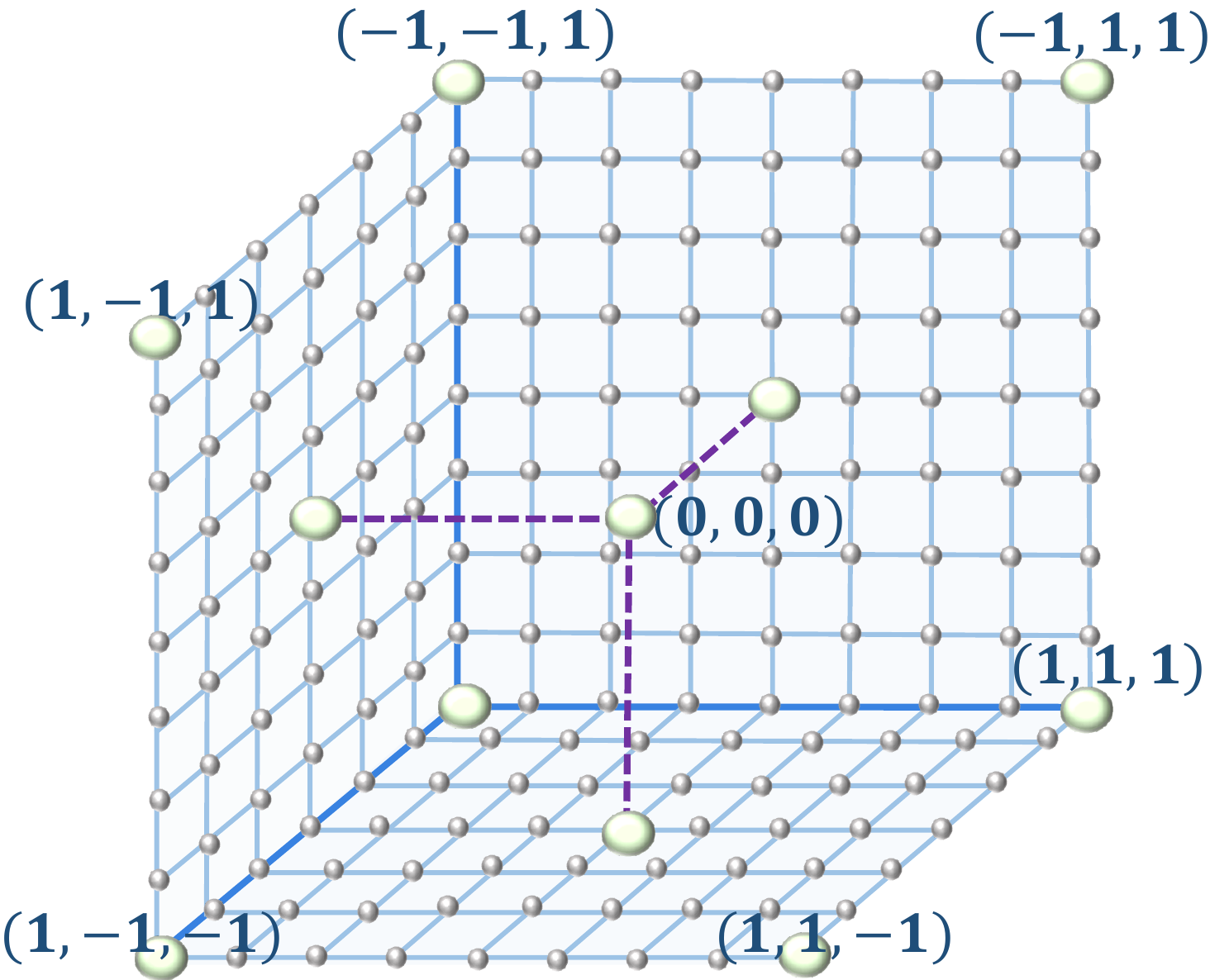}
  \end{center}
  \caption{The spatial anchors provided by $e_{csg}$.}
  \label{fig:csg}
\end{wrapfigure}
We observe that,
compared to existing 3D category-level models,
learning the internal distribution with spatial-invariant and receptive field-limited convolutional networks leads to more difficulties in producing plausible 3D structures.
%
Following~\cite{csg_singan}, 
which alleviates a similar issue in image generation by introducing spatial inductive bias,
we introduce spatial inductive bias into our pipeline by lifting the normalized Cartesian Spatial Grid (CSG) to 3D:
\begin{equation}
  e_{csg}(x, y, z) = 2 \cdot [\frac{x}{W} - \frac{1}{2}, \frac{y}{H} - \frac{1}{2}, \frac{z}{U} - \frac{1}{2}],
  \end{equation}
where $W$, $H$ and $U$ are size of the volume along the $x-$, $y-$ and $z-$axis. 
As illustrated in Figure~\ref{fig:csg}, the grid is equipped with distinct spatial anchors, empowering the model with better spatial localization.
The spatial anchors provided by $e_{csg}$ are injected into the noise volume $z_1$ at the coarsest level:
$\volgen_1 = G_1(z_1, e_{csg}).$
Note we only inject the spatial inductive bias at the coarsest scale, 
as the positional-encoded information will be propagated through subsequent scales by convolution operations~\citep{csg_singan}.
The effective receptive field of the generator and discriminator at this scale is around $40\%$ of the volume of interest and the image size, thus $G_1$ learns to generate the overall layout and structure of the objects in the scene.

Subsequently, $G_\scale$ at finer scales ($1<\scale<N$) learns to add details missing from previous scales.
Hence, each generator $G_\scale$ takes as input a spatial noise volume $z_\scale$ and an upsampled volume of $(\volgen_{\scale-1}) \uparrow^{\theta}$ output from the preceding scale. 
Specially, prior to being fed into $G_\scale$, $z_\scale$ is added to $(\volgen_{\scale-1}) \uparrow^{\theta}$,
and, akin to residual learning~\cite{residual}, $G_\scale$ only learns to generate missing details
:
\begin{equation} 
    \begin{gathered}
        \volgen_\scale = (\volgen_{\scale-1}) \uparrow^{\theta} + \; G_\scale \big( z_\scale, (\volgen_{\scale-1}) \uparrow^{\theta} \big),\; 1 < \scale < N.
    \end{gathered}
\end{equation}

At the finest scale, 
a 2D convolutional generator $G_N$ takes as input only the rendering $\imgen_{N-1}$ produced from the preceding scale,
and outputs a super-resolved image $\imgen_N$ with enhanced details:
$\imgen_N = G_N (\imgen_{N-1}).$
$G_N$ employs upsampling layers introduced in~\cite{stylenerf}, and produces the final image that twice the resolution of $\imgen_{N-1}$. To better guarantee the view consistency, we adopt a joint discrimination strategy as in \cite{eg3d} for the discriminator $D_N^{2D}$. To achieve a progressive learning of the internal distributions at different scales, the receptive field of discriminators $D_{\scale}^{2D}$ is limited so that the fineness of details in the generated scenes is gradually improved.

\subsection{Training loss}
The multi-scale architecture is sequentially trained, from the coarsest to the finest scale. 
We construct a pyramid of resized input observations, 
$\imset_\scale = \{ x_{\scale} \}_{\scale=1}^{N}$, 
for providing supervisions.
The GANs at coarser scales are frozen once trained.
The training objective is as followed:
\begin{equation}\label{eq:loss_total} 
    \begin{gathered}
        \min_{G_\scale} \max_{D_\scale^{2D}}  \loss_{adv}(G_\scale, D_\scale^{2D}) + \loss_{rec}(G_\scale) + \mathds{1}(\scale=N) \cdot \loss_{sw}(G_\scale).
    \end{gathered}
\end{equation}
where
$\loss_{adv}$ is an adversarial term,
$\loss_{rec}$ is a reconstruction term as similar in ~\cite{singan},
$\mathds{1}(\cdot)$ is an indicator function to activate the associated term \emph{iff} the condition is satisfied, 
$\loss_{sw}$ calculates the Sliced Wasserstein Distance (SWD) as in \cite{swd}.
SWD measures the distance between the textural distributions of two images, 
while neglecting the difference of the global layouts. 
Concretely, $\loss_{sw}$ is given by:
$\loss_{sw} = \loss_{sw}(\im_N, \imgen_N) =\sum_{k=1}^{K} \loss_{sw}(\tilde{f}^k, f^k)$,
where $\tilde{f}^k$ and $f^k$ are features from layer $k$ of a pre-trained VGG-19 network \citep{vgg}.
Please refer to~\cite{swd} for more details.
In the following,
we elaborate the designs of $\loss_{adv}$ and $\loss_{rec}$.

\paragraph{Adversarial loss} 
We use the WGAN-GP loss \citep{wgangp} as $\loss_{adv}$ 
for stabilizing the training. 
The discrimination score is obtained by averaging over the patch discrimination map of $D_{\scale}^{2D}$. 
Similar to~\cite{unconstrained}, we also condition the discriminators on the depth, of which the real samples $d_n$ can be derived with multi-view geometry techniques trivially, to help improve the structural plausibility of the generation.
During training,
we render the depth from the generated volume, and concatenate the depth and color images for the input to $\{D_{\scale}^{2D}\}_{n=1}^{N-1}$.
%
Note we do not use the depth discrimination for the 2D discriminator $D_{N}^{2D}$, 
and adopt the joint discriminator strategy as in \cite{eg3d} by concatenating the naively super-resolved $(\imgen_{N-1})\uparrow^{\mu_s}$ with the output from $G_{N}$. 
For preparing the real input to  discriminator $D_{N}^{2D}$, we upsample the resized observation $\im_{N-1}$ via bilinear upsampling and concatenate the upsampled image with the ground truth observation $x_N$.

\paragraph{Reconstruction loss}
Inspired by \cite{singan}, 
we introduce a specific set of input noise volumes to ensure that the noise within is able to reconstruct the underlying scene depicted in the observations $\imset$. 
Specifically, a set of fixed noise volume is defined as $\{z_{\scale}^{*}\}_{\scale=1}^{N-1} = \{z_{1}^{*}, 0, ..., 0\}$. 
The reconstructed radiance volumes and associated renderings are denoted as $\{V_{\scale}^{*}\}_{\scale=1}^{N-1}$ and $\{\imgen_{\scale}^{*}, \depgen_{\scale}^{*}\}_{\scale=1}^{N}$, respectively. 
Then the reconstruction loss $\loss_{rec}$ is defined as:
\begin{equation}\label{eq:loss_rec}
\loss_{rec} = \lambda_c||\imgen_{\scale}^{*}-\im_{\scale}||_2^2 + \mathds{1}(\scale < N) \cdot \lambda_d||\depgen_{\scale}^{*}-\depth_{\scale}||_2^2,
\end{equation}
where $\lambda_c$ and $\lambda_d$ are balance parameters and we set $\lambda_c=10$ and $\lambda_d=30$. As shown in Equation \ref{eq:loss_rec}, we use supervisions on both color images and depth images for achieving higher quality. Note that the depth penalty term is removed at the last scale since $G_N$ only works in the color image domain.


\section{EXPERIMENTS} \label{sec:experiment}

\setlength{\tabcolsep}{1pt}
\renewcommand{\arraystretch}{1}
\begin{figure*}[t]
\begin{center}
\begin{tabular}{cc}
\vspace{-0.4mm}
\makebox[0.25\textwidth][c]{\small{Training scene}}
 & \small{Random generation} \\
\multicolumn{2}{c}{\includegraphics[width=0.98\linewidth]{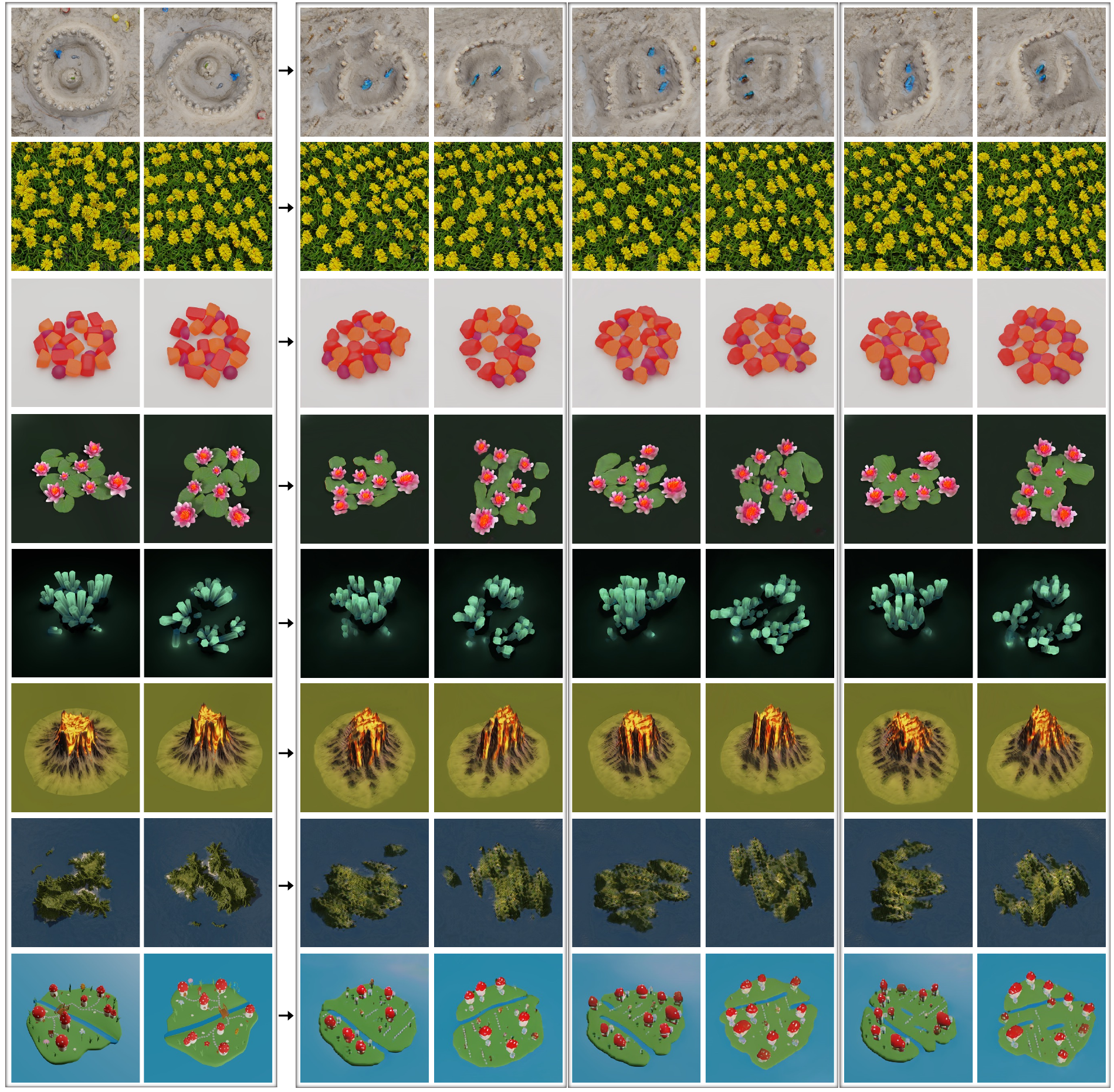}}\\
\end{tabular}
\end{center}
\caption{
Random scene generation.
After training on multi-view observations of an input scene,
\name learns to generate similar scenes with new objects and configurations.
At each row, we show the input scene (left) and three diverse generated scenes (right) under the same two viewpoints.
}
\label{fig:demos}
\end{figure*}
In the following,
we explain experimental settings, and present evaluation results.
More details and results can be found in the appendix~\ref{appendix} and the supplementary video.

\textbf{Data.}
We collected observation images from a dozen of synthetic and real scenes, 
exhibiting ample variations over the global arrangements and constituents, 
for evaluating our method. 
%
For each scene, 
we assume the volume of interest is within a unit cube,
and 200 observation images fully covering the scene are collected with cameras randomly distributed on a hemisphere.
%
%
Random natural scene generation of our method is demonstrated upon all collected scenes, whereas more evaluations are conducted on a subset (\emph{Stonehenge}, \emph{grass and flowers}, and \emph{island}).

\textbf{Evaluation measures.} 
We extend common metrics in single image generation to quantitatively assess $m=50$ scenes generated from each input scene.
For each generated scene, $k=40$ images at random viewpoints are rendered for evaluation under a multi-view setting:
a) \emph{SIFID-MV} 
measures how well the model captures the internal statistics of the input by SIFID~\citep{singan} averaged over multi-view images of a generated scene;
b) \emph{Diversity-MV} 
measures the diversity of generated scenes by the averaged image diversity~\citep{singan} over multiple views. 

\textbf{Neural scene variations.}
\label{para:variation} 
Figure \ref{fig:teaser} and \ref{fig:demos} presents qualitative results of \name,
where
the generated scenes depict reasonably new global layouts, and objects with various shapes
Note the observation images of the \emph{beach castle} at the top row are taken from a real-world scene,
and novel beach castles with various layouts are generated by \name.
and realistic looking.
These results suggest the efficacy of \name on modeling the internal patch distribution within the input scene. 
On \emph{grass and flowers}, 
which exhibits uniform yet complicated geometries and textures over an open field, 
\name produces high-quality random generation results. 
Moreover,
\name is able to capture the global illumination to some extent, as evidenced by the shadows around stones and islands, along with the illumination changes under spinning cameras on samples of \emph{mushroom}.
The supplementary video presents more visual results under spinning cameras.

\begin{figure*}[t!]
\begin{center}
\begin{tabular}{ccccc}
\vspace{-0.4mm}
\makebox[0.1\textwidth][c]{\small{Training}} &\multicolumn{4}{c}{\small{Random generation}} \\
\multicolumn{5}{c}{\includegraphics[width=0.985\linewidth]{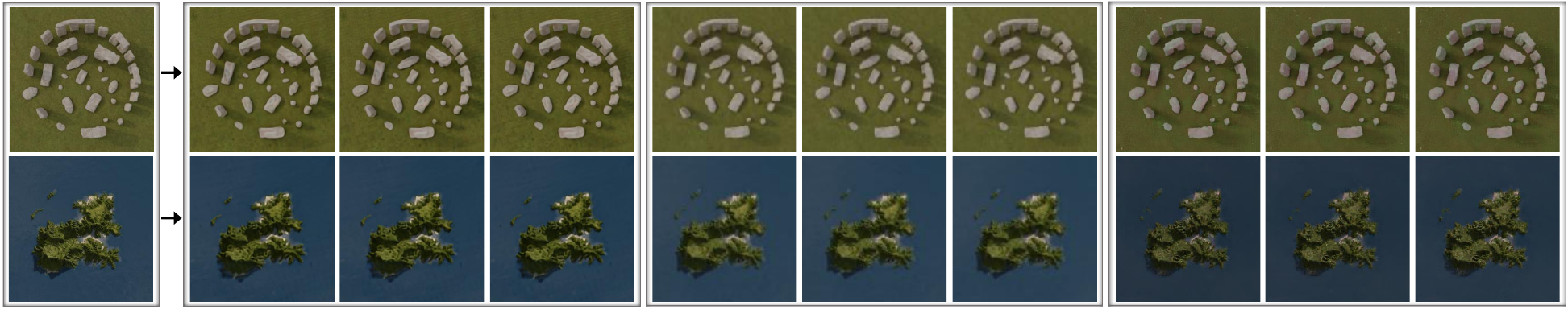}}\\
\vspace{-2mm}
\makebox[0.1\textwidth][c]{}& \makebox[0.02\textwidth][c]{}& 
\makebox[0.27\textwidth][c]{\small{GRAF}} & \makebox[0.29\textwidth][c]{\small{pi-GAN}} &
\makebox[0.27\textwidth][c]{\small{GIRAFFE}}
\\
\end{tabular}
\end{center}
\caption{
Qualitative results of baselines.
All baselines encounter severe mode-collapse issues.
}
\label{fig:comparison}
\end{figure*}

\textbf{Comparisons.}
\label{para:comparisons}
We compare to three state-of-the-art neural scene generative models, namely, GRAF~\cite{graf}, pi-GAN~\cite{pi-gan}, and GIRAFFE~\cite{giraffe}. 
We use their official codes for training these baselines.
Training details of each baseline can be found in the supplementary. 
For fair comparisons,
similar to \name,
we use ground truth cameras when training baselines,
instead of using random cameras from a predefined camera distribution.
The quantitative and qualitative results are presented in Table \ref{tab:comparison} and Figure \ref{fig:comparison}, respectively.
Table \ref{tab:comparison} shows that pi-GAN produces the best \emph{SIFID-MV} score, 
but the value of \emph{Diversity-MV} drastically degrades.
GRAF and GIRAFFE also exhibit a significantly degraded diversity score. 
The qualitative results in Figure \ref{fig:comparison} show that these baselines suffer from severe mode collapse,
due to the lack of diverse scene samples for learning category-level priors.

\begin{figure*}[t!]
\begin{center}
\begin{tabular}{cccc}
\makebox[0.13\textwidth][c]{\small{Input scene}} &\makebox[0.05\textwidth][c]{} & \multicolumn{2}{c}{\small{Random generation}} \\
\multicolumn{4}{c}{\includegraphics[width=0.985\linewidth]{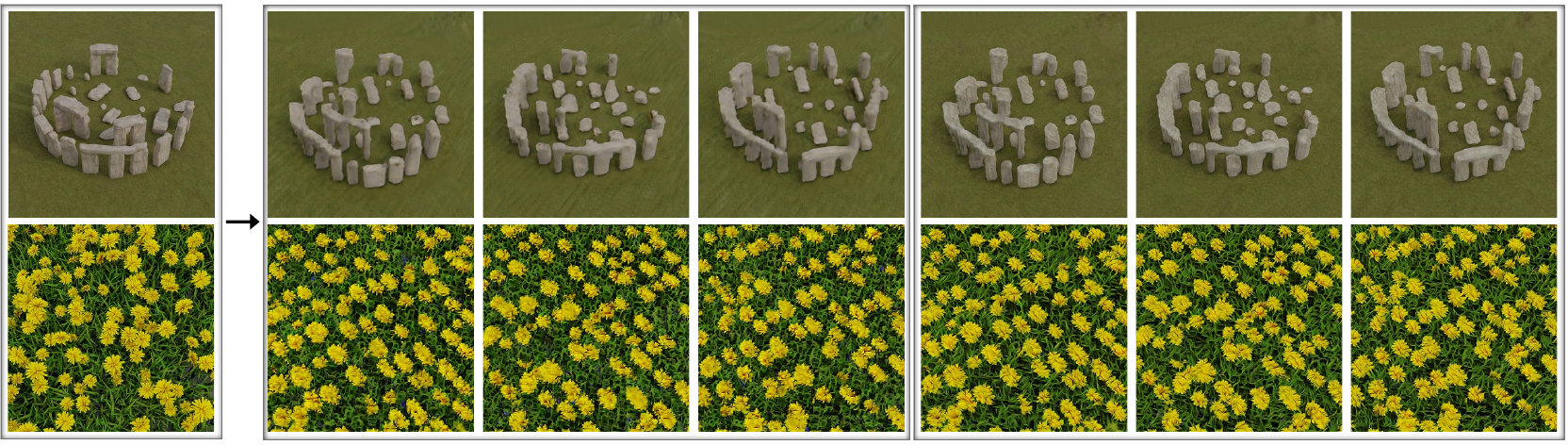}}\\
\makebox[0.13\textwidth][c]{}& \makebox[0.05\textwidth][c]{}&  
\makebox[0.38\textwidth][c]{\small{SinGRAV (wo. SWD)}} & \makebox[0.38\textwidth][c]{\small{SinGRAV}}\\
\end{tabular}
\end{center}
\caption{
\name vs. \name (wo. SWD).
Left: 
Without the SWD loss at the finest scale,
\name (wo. SWD) produces blurry textures and undesired artifacts.
Right:
Training with SWD loss significantly improves the texture quality.
}
\label{fig:ablation:wo_swd}
\end{figure*}

\setlength{\tabcolsep}{1pt}
\renewcommand{\arraystretch}{1}
\textbf{Validation of design choices.} 
\label{para:ablations}
We conduct experiments to evaluate several key design choices:

\emph{SWD loss.}
If $\loss_{swd}$ is eliminated, 
the internal distribution of generated scenes would differ greatly from that of the input,
leading to significantly increased \emph{SIFID-MV} at the first row in Table~\ref{tab:ablation}.
Correspondingly, 
the visual results in Figure~\ref{fig:ablation:wo_swd} also show that \name(wo.$\loss_{swd}$) produces blurry and less realistic textures, 
suggesting the efficacy of $\loss_{swd}$ in improving the visual quality.

\setlength{\tabcolsep}{1pt}
\renewcommand{\arraystretch}{1}

\begin{figure*}[h]
\begin{center}
\begin{tabular}{c}
\vspace{-0.4mm}
\includegraphics[width=.97\linewidth]{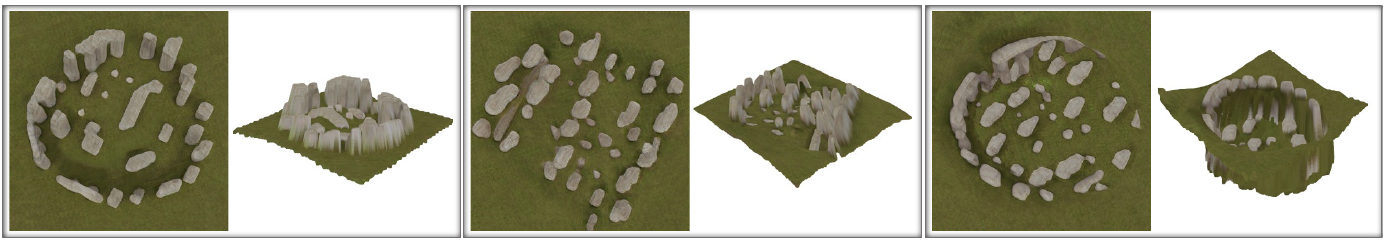}\\
\makebox[0.33\textwidth][c]{\small{\name}}   \makebox[0.3\textwidth][c]{\small{\name (wo. CSG)}} 
\makebox[0.33\textwidth][c]{\small{\name (wo. depth. sup.)}}
\\
\vspace{-4mm}
\end{tabular}
\end{center}
\caption{
Influence of spatial inductive bias and depth supervision. 
One generated sample with the corresponding point cloud is shown for each variant. 
Without the inductive bias (middle) or depth supervision (right), the generated scenes
exhibit implausible geometric structures,
while \name (left) preserves the spatial arrangements well.
}
\label{fig:ablation:csg_depth}
\end{figure*}

\emph{Spatial inductive bias.}
We build a variant \--- \name (wo.CSG), which is trained without the spatial anchor volume $e_{csg}$.
As shown in Table \ref{tab:ablation}, 
compared to \name,
\name (wo.CSG) produces a worse \emph{SIFID-MV} score (increases by $17\%$) and an increased \emph{Diversity-MV} score. While the latter 
\newpage
\noindent
\begin{minipage}[c]{0.5\textwidth}
\footnotesize
  \centering
  \setlength{\tabcolsep}{3pt}
  \captionof{table}{
  Quantitative comparisons.
  Top two on each metric are highlighted.
  All baselines suffer from severe mode collapse, producing low scores on Diversity-MV.
  }
    \begin{tabular}{lccc}
    \toprule
    Method & Img. res. & SIFID-MV $\downarrow$ & Diversity-MV $\uparrow$ \\
    \midrule
    GRAF    & $320^2$ & 0.4447 & 0.1337 \\
    pi-GAN  & $128^2$ & \textbf{0.0133} & 0.1157 \\
    GIRAFFE  & $320^2$ & 0.4710      & \textbf{0.3198} \\
    
    \name  & $320^2$ & \textbf{0.1113} & \textbf{0.7769} \\
    \bottomrule
    \vspace{1mm}
    \end{tabular}%
    \label{tab:comparison}%
\end{minipage}
\hfill
\begin{minipage}[c]{0.45\textwidth}
\scriptsize
  \centering
  \setlength{\tabcolsep}{1pt}
  \captionof{table}{Numerical results for variants of \name.
  }
  \begin{tabular}{lccc}
    \toprule
    Variant & Img. res. & SIFID-MV $\downarrow$ & Diversity-MV $\uparrow$ \\
    \midrule
    (wo. SWD) & $320^2$ & 0.2713 & 0.7730 \\
    (wo. CSG) & $320^2$ & 0.1307 & \textbf{0.8434} \\
    (wo. depth sup.) & $320^2$ & 0.1157 & 0.7910 \\
    (NeRF-depth) & $320^2$ & 0.1290 & 0.8046 \\
    (self-depth) &   $320^2$ & 0.1120 & 0.7862 \\
    (MLP) & $160^2$ & 0.1843 & 0.5235 \\
    (MLP-LLG) & $108^2$ & 0.2013 & 0.4621 \\
     \name  & $320^2$ & \textbf{0.1113} & 0.7769 \\ 
    \bottomrule
    \vspace{1mm}
    \end{tabular}%
    \label{tab:ablation}%
    
\end{minipage}

suggests increased diversity, 
we observe from the visual results that
the spatial arrangements of objects deviate significantly from that of the input, 
producing floating stones, as shown in the middle of Figure~\ref{fig:ablation:csg_depth}. 

\emph{Depth supervision}. 
The third row in~\ref{tab:ablation} presents numerical results of a variant \name (wo. depth sup.), which is trained without depth supervision.
\name (wo. depth sup.) achieves comparable numerical performance to \name. 
{
But, 
the extracted point clouds in Figure \ref{fig:ablation:csg_depth} reveal
implausible geometric structures in the generated scenes of \name (wo. depth sup.),
showing the importance of depth supervision.
%
%
On the other hand,
we also show that our framework can work with depth data obtained from various sources, in addition to rather clean depth map in our default setting. 
Specifically, 
we designed \name (NeRF-depth),
which uses the depth obtained from a NeRF~ \cite{nerf} reconstruction of the input scene,
and \name (self-depth),
which uses the depth obtained from the reconstructed volume with $z^*$ and $\lambda_d = 0$ in our framework.
Table \ref{tab:ablation} shows that \name (NeRF-depth) and \name (self-depth) are able to achieve comparable performance to \name, implying that \name is robust to the quality of the depth data.
That said, 
all these depth data of different quality provide sufficient supervision for learning global geometric structures.
}

\begin{figure*}[t!]
\begin{center}
\begin{tabular}{cccc}
\vspace{-0.4mm}
\makebox[0.14\textwidth][c]{\small{Input scene}} & \multicolumn{3}{c}{\small{Random generation}} \\
\multicolumn{4}{c}{\includegraphics[width=0.985\linewidth]{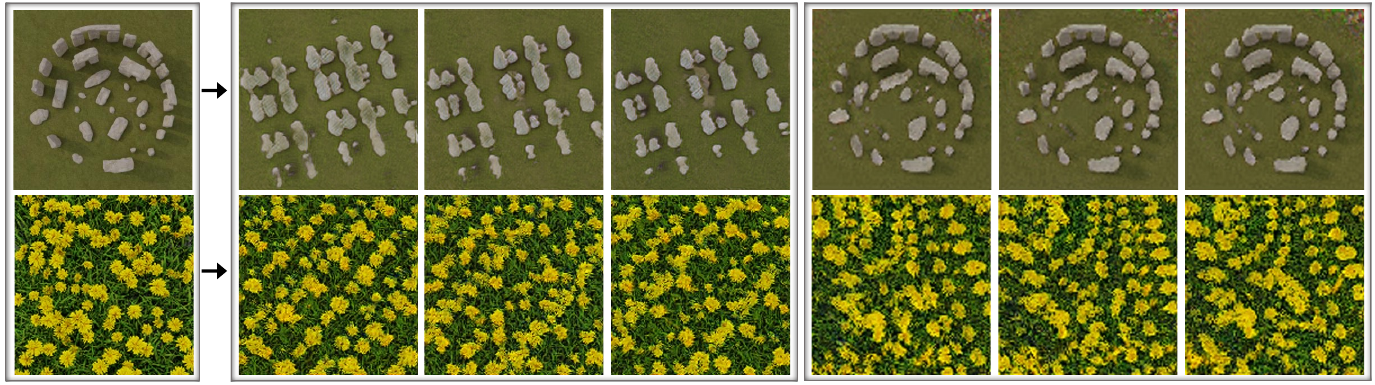}}\\
\vspace{-0.4mm}
\makebox[0.14\textwidth][c]{}& \makebox[0.03\textwidth][c]{}&  
\makebox[0.4\textwidth][c]{\small{\name (MLP)}} & \makebox[0.4\textwidth][c]{\small{\name (MLP-LLG)}}\\
\vspace{-4mm}
\end{tabular}
\end{center}
\caption{
Qualitative results of using MLP-based representations. 
\name (MLP) with globally conditioned MLPs produces grid patterns, which are alleviated by the use of local latent grid in \name (MLP-LLG). 
Nevertheless,
both variants suffer from severe mode collapse,
generating almost identical scene samples.
}
\vspace{-0.2cm}
\label{fig:ablation:nerf}
\end{figure*}

\emph{MLP vs. voxel}.
The numerical results of \name (w. MLP), which uses a conditional MLP-based radiance field as in GRAF, are reported in Table \ref{tab:ablation}.
The highest image resolution is $160\times160$ due to overly high memory consumption. 
\name (w. MLP) degrades in both the quality and diversity, which is also reflected by the visuals in Figure \ref{fig:ablation:nerf}. 
Note that, although the same patch discrimination strategy is used, \name (w. MLP) suffers from severe mode collapse.
We believe this is because
the generative output of coordinate-based MLPs is very likely to be dominated by the input coordinates.
The poor generation is also possibly a product of the conflict between the lack of locality in the fully connected layers~\cite{repmlp} and the limited receptive field in the adversarial training. 
In addition, we train a variant \name (w.MLP-LLG), which incorporates a local latent grid proposed in \cite{unconstrained} to increase the capacity of the MLP-based representation. 
Figure \ref{fig:ablation:nerf} shows that the generated layouts are improved with the increased locality, however, severe mode collapse still exists.
On the other hand, 
we believe more effort can be made in the future to adapt MLP-based representations for our task.

\begin{figure*}[h]
\begin{center}
\begin{tabular}{cccccc}
\vspace{-0.4mm}
\makebox[0.12\textwidth][c]{\:\:Input scene}& \makebox[0.02\textwidth][c]{}& 
\multicolumn{4}{c}{\small{Random generation}} \\
\multicolumn{6}{c}{\includegraphics[width=0.99\linewidth]{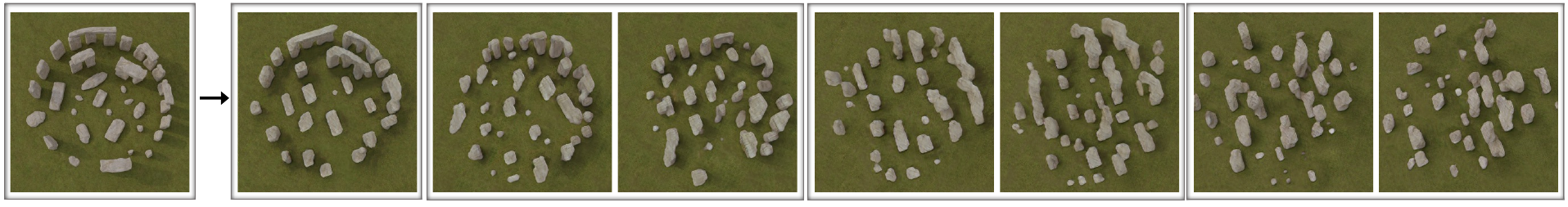}}\\
\vspace{-3mm}
\makebox[0.12\textwidth][c]{}& \makebox[0.02\textwidth][c]{}& 
\makebox[0.12\textwidth][c]{\small{$N=6$}}& 
\makebox[0.23\textwidth][c]{\small{$N=5$}} & \makebox[0.23\textwidth][c]{\small{$N=4$}} &
\makebox[0.23\textwidth][c]{\small{$N=3$}}
\\
\vspace{-2mm}
\end{tabular}
\end{center}
\caption{
Influence of different number of scales. 
Training with more scales is beneficial to the modeling of global arrangements, while a model with less scales tends to capture only local textures.
}
\label{fig:ablation:vary_scale}
\end{figure*}

\setlength{\tabcolsep}{1pt}
\renewcommand{\arraystretch}{1}
\emph{Influence of varying pyramid depth.}
We train variants with various numbers of scales.
Specifically, for training a variant with $t$ ($t < N$) scales, we use generators $\{G_{\scale}\}_{\scale=N-t+1}^N$ and discriminators $\{D_{\scale}^{2D}\}_{\scale=N-t+1}^N$ to preserve the final image resolution. 
As shown in Figure~\ref{fig:ablation:vary_scale},
with less scales, the effective receptive field at the coarsest scale is rather small, resulting in a model that only captures local properties,
whereas,
more scales allows modeling plausible global arrangements.

\textbf{Applications.}
We show the application potential of \name in common 3D modeling process. 
Specifically, we derive three applications with \name,
including 3D scene editing with removal and duplicate operations, composition, and animation. 
Figure~\ref{fig:teaser} presents the results and more can be found in the supplementary video.
\renewcommand{\arraystretch}{1}
\section{CONCLUSION}
In this work, we make a first attempt to learn a deep generative neural scene model from visual observations of a single scene.
Once trained on a single scene,
the model can generate novel scenes with plausible geometries and arrangements, that can be rendered with pleasing viewing effects.
The importance of key design choices are validated.
Despite successful demonstrations, our technique has a few limitations of its current designs.

While our model learns from a single scene,
bypassing the need for collecting data from many homogeneous 3D samples,
a rather complete 3D scene is yet required to obtain the multi-view data for training.
%
Moreover,
albeit validated,
the use of voxel grids inherently limits the network ability in modeling fine details, 
consequently hindering the model from achieving high-resolution imagery. 
Our remedy is to incorporate a 2D neural renderer that operates on 2D domain to super-resolve the imagery,
\revise{
which inevitably introduces the multi-view inconsistency. 
There are view inconsistencies on complex textural areas, such as the thin structures in the Grass and Flowers scene.}
A future direction would be to overturn this design, with more endeavor on exploiting MLP-based representations to model continuous volumes.
Besides,
the proposed method, in its current form,
sometimes produces artifacts in the background,
as it does not incorporate special designs for modeling the background.
Hence,
it would also be worth addressing such an issue, especially for scenes with complicated background, potentially with designs inspired from other generative neural scene models. 

\bibliography{ref}
\bibliographystyle{plain}


\newpage 

\section{Appendices}
\label{appendix}
\revise{In this section, we provide additional implementation details in Section \ref{sec:implementation}, additional experimental results in Section \ref{sec:more_exp} and list outlines in Section \ref{sec:video_results} for video demonstrations provided in the supplementary video.}

\subsection{Implementation details} \label{sec:implementation}
We describe more implementation details including detailed network structures for \name in Section \ref{sec:net_design}, details for data creation in Section \ref{sec:data_creation}, details for evaluation measures in Section \ref{sec:metrics} and details for demonstrated applications in \ref{sec:application}. In addition, we provide the detailed training settings for \name in Section \ref{sec:training} and training details for three baselines in Section \ref{sec:training_baseline}.

\subsubsection{More details for network design} \label{sec:net_design}
\revise{\paragraph{Overview}
All 3D generators $\{G_{\scale}\}_{\scale=1}^{N-1}$ share the same structure, 
with the same number of 3D convolutional layers (kernel size 3 and stride 1). 
The number of layers is set to $5$ (small receptive field for small objects in the scene), or $7$ (large receptive field for large objects), according to different scene constituents.
All 2D discriminators $\{D_{\scale}^{2D}\}_{\scale=1}^{N}$ have the same number (5 or 7) of 2D convolutional layers (kernel size 3 and stride 1). 
The resolution of the volume and the image start from $40\times40\times40$ and $32\times32\times32$, and increase by a factor of $\theta=4/3$ and $\mu_r=1.5$, respectively.  
We set $N=6$ by default, 
and at scale $N-1$, 
the volume and image resolution reach $126^3$ and $160^2$ separately. 
The above setting is used on all data, except the \emph{grass and flows}, and \emph{island} that possess rather fine-grained texture details.}


\paragraph{Detailed structure for $\{G_{\scale}\}_{\scale=1}^{N-1}$}
For generators ranging from the coarsest scale to scale $N-1$, we adopt the same structure. Specifically, for generators $\{G_{\scale}\}_{\scale=1}^{N-1}$, we adopt an $L$-layer structure, which is designed with 3D convolutional layers, Batch Normalization and LeakyReLU activation. The detailed structure of  $\{G_{\scale}\}_{\scale=1}^{N-1}$ is summarized in Table \ref{tab:generator}. As shown in Table \ref{tab:generator}, the number of input channels, $\text{IC}_{\scale}^{V}$, is adjusted according to the difference of the input volume for different scales. At the coarsest level, to make the volume adapt to the spatial anchor volume $e_{csg}$, we set $\text{IC}_{1}^{V}=3$. For the remaining 3D generators, we set $\text{IC}_{\scale}^{V}=4$. In addition, for generators whose scale is in range $[2, N-1]$, a residual connection is set up between the output volume and the upsampled $\tilde{V}_{\scale-1}$, \emph{i.e}, the generators $\{G_{\scale}\}_{\scale=2}^{N-1}$ learn to add more details to the synthesized scene.

\setlength{\tabcolsep}{1.8pt}
\renewcommand{\arraystretch}{1}
\begin{table*}[htp]
   \caption{Detailed structure of the generators $\{G_{\scale}\}_{\scale=1}^{N-1}$. Specifically, ``conv3d", ``batchNorm3d" represent the 3D convolution layer and batch normalization layer respectively. $\text{IC}_{\scale}^{V}$ denotes the number of channels within the input volume. We set $\text{IC}_{\scale}^{V}=3$ for $\scale=1$ and $\text{IC}_{\scale}^{V}=4$ for $\scale>1$. $L$ is set to $5$ or $7$. }
    \label{tab:generator}
   \begin{center}
  \begin{tabular}{|c|l|ccccc|}
      \hline
    Index & Name & Kernel & Stride &Pad & Input ch. & Output ch. \\
    \hline
    $1$ & Conv3d + BatchNorm3d + LeakyReLU (0.2) & $3\!\times\!3\!\times\!3$ & 1 & 1 & $\text{IC}_{\scale}^{V}$ & $32$ \\
    
    $2$ & Conv3d + BatchNorm3d + LeakyReLU (0.2) & $3\!\times\!3\!\times\!3$ & 1 & 1 & $32$ & $32$ \\
    
    $[3,{L-1}]$ & Conv3d + BatchNorm3d + LeakyReLU (0.2) & $3\!\times\!3\!\times\!3$ & 1 & 1 & $32$ & $32$\\
    
    $L$ & Conv3d & $3\!\times\!3\!\times\!3$ & 1 & 1 & $32$ & $4$\\
    \hline
    \end{tabular}
  \end{center}
\end{table*}

\paragraph{Structure of $G_N$}
At the finest level, we adopt a 2D neural generator $G_N$ to directly lift the rendered images to the output resolution $320\!\times\!320$. We adopt an upsampling layer proposed in \cite{stylenerf} at the first layer, after which we design $4$ layers with 2D convolutional layers and Instance Normalization. The detailed structure is given in Table \ref{tab:upsampler}. Besides, we utilize a residual connection between the final output and the resized input image, \emph{i.e.}, $\tilde{x}_{N-1}\uparrow$.

\setlength{\tabcolsep}{1.5pt}
\renewcommand{\arraystretch}{1}
\begin{table*}[htp]
   \caption{Detailed structure of the generators $G_N$. Specifically, "conv2d" represent the 2D convolution layer. "InsNorm2d" denotes Instance Normalization.}
    \label{tab:upsampler}
   \begin{center}
  \begin{tabular}{|c|l|cccc|cc|}
      \hline
    Index & Name & Kernel & Stride &Pad &I/O & Input res & Output res\\
    \hline
    $1$ & Upsampling layer & - & - & - & $3/3$ & $160\!\times\!160$ &  $320\!\times\!320$\\
    
    $2$ & Conv2d + InsNorm2d + LeakyReLU (0.2) & $3\!\times\!3$ & 1 & 1 & $32/16$ & $320\!\times\!320$ &  $320\!\times\!320$\\
    
    $3$ & Conv2d + InsNorm2d + LeakyReLU (0.2) & $3\!\times\!3$ & 1 & 1 & $16/8$ & $320\!\times\!320$ &  $320\!\times\!320$\\
    
    $4$ & Conv2d + InsNorm2d + LeakyReLU (0.2) & $3\!\times\!3$ & 1 & 1 & $8/8$ & $320\!\times\!320$ &  $320\!\times\!320$\\
    
    $5$ & Conv2d  & $3\!\times\!3$ & 1 & 1 & $8/3$ & $320\!\times\!320$ &  $320\!\times\!320$\\
    
    \hline
    \end{tabular}
  \end{center}
\end{table*}
\setlength{\tabcolsep}{2.5pt}
\renewcommand{\arraystretch}{1}
\begin{table*}[htp]
   \caption{Detailed structure of the generators $\{G_{\scale}\}_{\scale=1}^{N-1}$. Specifically, ``conv2d" represent the 2D convolution layer. $\text{IC}_{\scale}^{I}$ denotes the number of channels within the input for $D_{\scale}^{2D}$. We set $\text{IC}_{\scale}^{I}=4$ for $\scale<N$ and $\text{IC}_{\scale}^{V}=6$ for $\scale=N$. $L$ is set to $5$ or $7$.}
    \label{tab:discriminator}
   \begin{center}
  \begin{tabular}{|c|l|ccccc|}
      \hline
    Index & Name & Kernel & Stride &Pad & Input ch. & Output ch.\\
    \hline
    $1$ & Conv2d + LeakyReLU (0.2) & $3\!\times\!3$ & 1 & 0 & $\text{IC}_{\scale}^{I}$ & $32$\\
    
    $2$ & Conv2d + LeakyReLU (0.2) & $3\!\times\!3$ & 1 & 0 & $32$ & $32$\\
    
    $[3, L-1]$ & Conv2d + LeakyReLU (0.2) & $3\!\times\!3$ & 1 & 0 & $32$ & $32$ \\
    
    $L$ & Conv2d & $3\!\times\!3$ & 1 & 0 & $32$ & $1$  \\
    \hline
    \end{tabular}
  \end{center}
\end{table*}
\paragraph{Detailed structures for and $\{D_{\scale}^{2D}\}_{\scale=1}^{N}$} We design $\{D_{\scale}^{2D}\}_{\scale=1}^{N}$ in a similar manner to the design of $\{G_{\scale}\}_{\scale=1}^{N-1}$, \emph{i.e.}, all of $\{D_{\scale}^{2D}\}_{\scale=1}^{N}$ are designed with the same structure implemented with 2D convolutional layers and LeakyReLU() activation. In Table \ref{tab:discriminator}, we provide the detailed information of the structure. The discriminators have the same number of layers as the generators $\{G_{\scale}\}_{\scale=1}^{N-1}$. For the first layer of the generators $\{D_{\scale}^{2D}\}_{\scale=1}^{N-1}$, we set $\text{IC}_{1}^{I}=4$ since their input is formed by concatenating the rendered color image $\tilde{x}_{\scale}$ and depth map $\tilde{d}_{\scale}$. For the discriminator $D_{N}^{2D}$ at the finest scale, we set $\text{IC}_{N}^{I}=6$ as its input is the concatenation of $\tilde{x}_{N}$ and $\tilde{x}_{N-1}\uparrow$, which is the output from $G_N$ and the super-resolved image from the preceding scale respectively.

\subsubsection{Additional details for data creation} \label{sec:data_creation}
The 3D models are collected from \href{https://www.turbosquid.com}{this website}\footnote{https://www.turbosquid.com}, under \href{https://blog.turbosquid.com/turbosquid-3d-model-license/\#3d-model-license}{TurboSquid 3D Model License}\footnote{https://blog.turbosquid.com/turbosquid-3d-model-license/\#3d-model-license}.
During the process of data creation, we scale the scenes so that the volume of interest of them stay within a cube with side width=$2$ (within the range $[-1,1]$). 
To obtain the data for training our model,
on synthetic input scenes,
we utilize path tracing renderer in Blender to get the multi-view RGB-D observation. 
As for real-world scenes,
the visual images are collected from the internet (Google Earth and Sketchfab),
we then utilize COLMAP to reconstruct high-quality depth and camera poses, 
which is a common practice in many neural scene representation works.
Specifically,
for \emph{Stonehenge},\emph{grass and flowers}, \emph{island},\emph{mushroom} and \emph{water lily}, we use a camera with FOV=$33.40^{\circ}$ (focal length=76mm). For \emph{crystal}, \emph{candies} and \emph{volcano}, we use a camera with FOV=$49.13^{\circ}$ (focal length=50mm). 
The viewpoints are sampled randomly on a hemisphere with a radius of $3.5$.

\subsubsection{More details for evaluation metrics}\label{sec:metrics} 
In this section, we provide more details for calculating the scores for the evaluation metrics including \emph{SIFID-MV} and \emph{Diversity-MV}. Specifically, for each dataset, we randomly sample $M=40$ pairs of color image $x_m$ and camera pose $ps_m$, and $m$ denotes the index running over the sampled viewpoints. Then, we generate $J=50$ scenes, which are rendered under the sampled viewpoints to get $40$ rendered images $\tilde{x}_m^j$ per scene. Note that the notions for scales are omit here for brevity since we only use the rendered image at the finest scale for evaluation. Herein, we calculate the score for \emph{SIFID-MV} according to 
\begin{equation}\label{eq:sifid_mv}
    {\text{SIFID-MV}}=\frac{1}{M}\frac{1}{J}\sum_{m=1}^{M}\sum_{j=1}^{J} \text{SIFID}(\tilde{x}_m^j, x_m),
\end{equation}
where $\text{SIFID}(\cdot, \cdot)$ denotes the SIFID metric proposed in \cite{singan}, which is designed for measuring the distance between the internal distributions of two single images.

And we calculate the score for \emph{Diversity-MV} according to 
\begin{equation} \label{eq:diversity}
    {\text{Diversity-MV}}=\frac{1}{M} \sum_{m=1}^{M} \frac{\text{std}({\{\tilde{x}_m^{j}\}_{j=1}^{J}})}{\text{std}(x_m)},
\end{equation}
where $\text{std}(\cdot)$ denotes calculating the standard deviation of each pixel over all of the rendered images under the same viewpoints. As shown in Equation (\ref{eq:diversity}), before calculating the averages over all of the viewpoints, we normalize the score under single viewpoint with the standard deviation over all of the pixels within the corresponding input observation.

\subsubsection{Implementation details of applications} 
\label{sec:application}
In this section, 
we present the implementation details of the demonstrated applications, 
including \emph{Scene Animation} and \emph{Scene Editing}. 
Since the generator $G_N$ at the finest scale only operates in 2D rendering domain and functions mainly as a faithful super-resolving module, we omit it for brevity in the descriptions below.  

\paragraph{\textbf{Scene Animation.}}
To plausibly animate a generated 3D scene for $T$ consecutive time steps, we need to construct a sequence of radiance volumes $\{\tilde{V}_{N-1}^{(t)}\}_{t=1}^{t=T}$ with similar layouts but slightly changing content. 
In our implementation, we achieve this by utilizing the generators to produce $\{\tilde{V}_{N-1}^{(t)}\}_{t=1}^{t=T}$ from the modified versions of the pyramid of noise volumes $\{z_n\}_{n=1}^{n=N-1}$, which is used to generate the original scene $\tilde{V}_{N-1}$.
Specifically, we inject random changes smoothly for $T$ time steps $\{t\}_{t=1}^{T-1}$, and restrict the resulting noise volumes to stay close to the original ones $\{z_i\}_{n=1}^{n=N-1}$.
Formally,
the way we construct the random walk is given by:
\begin{equation}
\begin{split}
    z_{n}^{(t+1)} &= \alpha z_{n}^{(1)} + (1 - \alpha)(z_n^{(t)} + {\delta}_n^{(t+1)}), \\
    {\delta}_n^{(t+1)} &= \xi (z_n^{(t)} - z_n^{(t-1)}) + (1 - \xi){\mu}_{n}^{(t+1)},\\
\end{split}
\end{equation}
where $n$ denotes the index running over the pyramid of the noise volumes. 
$z_n^{(1)}$ is the noise volume at the first step and is set as the original noise volume at scale $n$, \emph{i.e.}, $z_n^{(1)}=z_n$. ${\delta}_{n}^{(t+1)}$ is the injected change at time step $t+1$, and ${\mu}_{n}^{(t)}$ is a noise volume that has the same distribution as $z_n$. 
$\alpha$ and $\xi$ are two parameters that can be tuned to achieve different animation effects.
Concretely, $\alpha$ controls the degree of preserving the original scene, \emph{i.e.}, the similarity between the original scene $\tilde{V}_{N-1}$ and the animated scene at time step $t+1$. 
$\xi$ controls the smoothness between the adjacent frames and the rate of introduced changes in the generated frame. 
Note that the noise volumes at the coarser scales are able to control the global layout of the generated scenes, on which slight modifications might drastically alter the scene. 
Thus we choose to inject the random changes in finer scales while keeping the noise volumes at the coarser scales unchanged for each time step.
For obtaining the demonstrated animation effects for \emph{candles} and \emph{crystal}, we let the random change injection start from the third scale and set $\alpha=0.58$ and $\xi=0.45$.

\paragraph{\textbf{Scene Editing.}} 
For editing a scene, 
we first need to localize the target area, 
then simply manipulate the volume $\tilde{V}_{N-1}$,
and finally downscale the edited volume and let the volume go through the pyramid of generators to harmonize the edited scene. 
We provide detailed description of these steps below.

\emph{Step1: object area localization.}
For finding an area of interest, 
we first extract the mesh from a scene volume $\tilde{V}_{N-1}$,
and interactively locate the bounding box of the area of interest in a 3D mesh visualizer, e.g., Blender in our prototype implementation. 
Then we construct a 3D mask $\mathcal{M}$,
where the area within the bounding box is filled with $1$ and the rest is set to $0$. 
Besides, we also sample a point within the area representing 'air' and use its color and density as $\mathbf{c}_{empty}$ and $\sigma_{empty}$.

\emph{Step2: volume manipulation.}

For the application \emph{Removal}, we manipulate the volume via 
\begin{equation}
\tilde{V}_{N-1}(\mathcal{M}=1)=(\mathbf{c}_{empty}, {\sigma}_{empty}).
\end{equation}

For the application \emph{Duplicate}, we select an area with empty voxels within the volume, which is masked by ${\mathcal{M}}_{empty}$. Then we set the empty area to the values extracted from the object area covered by $\mathcal{M}$. That is, we manipulate the volume via
\begin{equation}
\tilde{V}_{N-1}({\mathcal{M}}_{empty}=1)={\tilde{V}}_{N-1}(\mathcal{M}=1).
\end{equation}

For the application \emph{Move}, we achieve the effects by first conducting the \emph{Duplicate} operation and then performing the \emph{Removal} operation to $\tilde{V}_{N-1}$. 

For the application \emph{Composition}, we select the object of interest in $k$ generated scenes and construct a group of masks $\{{\mathcal{M}}_i\}_{i=1}^{k}$. 
Then we set $k$ target bounding boxes within an extra scene, which are destinations where we want to inject the new objects in. Consequently, we get $k$ target masks $\{{\mathcal{M}}_{i}^{dest}\}_{i=1}^{k}$.
Then we combine the objects into a single scene by putting the extracted sub-volume covered by ${\mathcal{M}}_i$ for each object in corresponding area ${\mathcal{M}}_i^{dest}$.

\emph{Step3: neural editing.}
After each kind of volume manipulation mentioned above, we downscale it to the size of $\tilde{V}_{3}$ and then let $\{G_{n}\}_{n=3}^{N-1}$ to generate the final volume $\tilde{\tilde{V}}_{N-1}$.
This step will smooth the discontinuities caused by the naive volume manipulations. 
Then we can generate the renderings from the newly generated scene.

\subsubsection{Training details for \name} \label{sec:training} 
\emph{Weight initialization.}
For the networks at the coarsest level, $\{G_1, D_1\}$, we adopt a random initialization. For generator $G_{\scale}$ and discriminator $D_{\scale}$ ($\scale>1$), we initialize the weights using the weights from trained networks of the preceding scale, \emph{i.e.}, weights of $G_{\scale-1}$ and $D_{\scale-1}$. We empirically find that using the weights from the pre-trained model of the preceding scale is helpful to stabilize the training.

\emph{Parameter settings.} The weight of gradient penalty term in the adversarial loss is set to $0.1$. We adopt the Adam optimizer for optimizing both the generator and discriminator, for both of which the learning rate is set to $0.0005$ and the momentum parameters are set as $\beta_1=0.9, \beta_2=0.999$.

\emph{Training.} We sequentially train the networks at different scales in a coarse-to-fine manner and the networks at coarser levels are fixed after being trained.
Fixing the networks at the coarser scales is helpful to stabilize the training process of the network at finer scales, since the results at the beginning of the training at each scale is messy.
At each scale $s$, we train the networks for $80$ epochs. Specifically, we first train $G_{\scale}$ using the reconstruction loss defined in Equation (\ref{eq:loss_rec}) for $20$ epochs and then train $\{G_{\scale}, D_{\scale}^{2D}\}$ together using the loss defined in Equation (\ref{eq:loss_total}). In each iteration, we alternate between 3 gradient steps for the discriminator $D_{\scale}^{2D}$ and 3 steps for the generator $G_{\scale}$.

\emph{Time and platform.} We train \name with $4$ NVIDIA Tesla V100 GPUs (32GB memory) for around $2$ days. For generating a new scene, the inference time is $0.32$ seconds and the memory consumption is about $5$ GB.

\paragraph{Parameter settings for different data} In Table \ref{tab:train_params}, we list the detailed parameter settings used for training \name on the collected datasets.  For most of the data, the settings shown in the first row of Table \ref{tab:train_params} are applicable. For training on \emph{island}, we slightly adjust the number of layers in $\{G_{\scale}\}_{\scale=1}^{N-1}$ and $\{D_{\scale}^{2D}\}_{\scale=1}^{N}$ to make them have a smaller receptive field, which is helpful to learn the fine textures existed in \emph{island} dataset. For training on \emph{water lily}, we slightly increase the resolution of $\tilde{V}_{1}$ considering the complicated geometries of the water lily. Besides, with the gradually increased resolution of volumes and rendered images, we adjust the batch size at each scale in order to accommodate the training to the memory budget of the hardwares. Note that the parameter tweaking for \emph{island} and \emph{water lily} does not bring significant gains, instead, it just slightly improve the results.

\begin{table}[htbp]
  \centering
  \small
  \caption{Training parameters for \name on different datasets. $L$ denotes the number of layers in $G_{\scale}$ or $D_{\scale}^{2D}$.}
  \vspace{4mm}
    \begin{tabular}{|c|c|c|c|c|l|l|}
    \hline
    Datasets & $L$ & \begin{tabular}[x]{@{}c@{}} Res. of \\ ${\tilde{V}}_1$ / ${\tilde{x}}_1$\end{tabular}    & \begin{tabular}[x]{@{}c@{}} Res. of \\ ${\tilde{V}}_{N-1}$ / ${\tilde{x}}_{N-1}$\end{tabular}   & \begin{tabular}[x]{@{}c@{}}Final\\img Res.\end{tabular} & \begin{tabular}[x]{@{}c@{}}batch size for\\$\tilde{x}_{\scale}$ at 6 scales\end{tabular} & \begin{tabular}[x]{@{}c@{}}batch size for\\$\tilde{x}_{\scale}^{*}$ at 6 scales\end{tabular}\\
    \hline
    \begin{tabular}[x]{@{}c@{}}\emph{Stonehenge}, \emph{grass and}\\\emph{flowers}, \emph{mushroom}, \emph{crystal}, \\ \emph{candies},\emph{volcano} \end{tabular} & 7     & $40\!/\!32$ & $126\!/\!160$ & $320$   & $[6,6,6,5,2,2]$ & $[2,2,2,1,1,1]$ \\
    \hline
    \emph{Island} & 5     & $40\!/\!25$ & $126\!/\!160$ & $320$   & $[6,6,6,6,3,2]$ & $[2,2,2,1,1,1]$ \\
    \hline
    \emph{Water lily} & 7     & $53\!/\!32$ & $156\!/\!160$ & $320$   & $[6,6,4,2,1,1]$ & $[2,2,2,1,1,1]$ \\
    \hline
    \end{tabular}%
  \label{tab:train_params}
\end{table}%

\emph{Sampling strategy for volume rendering}. In the rendering process, we adopt a uniform sampling strategy, \emph{i.e.}, we uniformly sample a number of points within the range of the \emph{near} point and \emph{far} point along each ray cast from each pixel. Considering the increased fineness of textures within the generated scene at different scales, we progressively enlarge the number of sampling. Specifically, the number of sampling points start from $64$ at scale $1$ and reaches $128$ at the finest scale.

\subsubsection{Detailed training settings for baselines} \label{sec:training_baseline}
For training the baseline methods, we keep most of the parameters the same as those provided in the exemplar configurations. The parameters we modified according to our settings are listed below:

\textbf{GRAF} \cite{graf}: On each dataset, we train more than $4500$ epochs until convergence. The patch size used during training is set to $32$. Due to the patch based training process, it supports to be trained at $320\times320$ resolution and convergences slower.

\textbf{pi-GAN} \cite{pi-gan}: We train $685$ epochs on each dataset and adopt the progressive training strategy as in \cite{pi-gan}. Specifically, we train $150$ epochs with batch size $=30$ at image resolution $32\times32$, $270$ epochs with batch size $=12$ at image resolution $64\times64$ and $265$ epochs with batch size $=8$ at image resolution $128\times128$.

\textbf{GIRAFFE} \cite{giraffe}: We set the parameters according to the instructions provided in the \href{https://github.com/autonomousvision/giraffe}{official implementation} \footnote{https://github.com/autonomousvision/giraffe}. Specifically, we set the number of object field to $1$ and use a background field. For training on our datasets, we separately set the parameters for the foreground field in the following manner:
\begin{itemize}
    \item \emph{Stonehenge}: The scaling range is set to $[0.45, 0.55]$ and the translation range is set to $[0.0, 0.0]$ for x-axis, y-axis and z-axis.
    \item \emph{Grass and flowers}:  The scaling range is set to $[0.7, 0.8]$ and the translation range is set to $[0.0, 0.0]$ for x-axis, y-axis and z-axis.
    \item \emph{island}: The scaling range is set to $[0.45, 0.55]$ for x-axis, y-axis and z-axis.  And the translation range for x-/y- axis is set to $[-0.15, 0.15]$, while the translation range for z-axis is set to $[0.0,0.0]$.
\end{itemize}
Besides, for training with GIRAFFE \cite{giraffe} method, we let the low-res image output from the volume rendering have a resolution of $20\times20$, so that the resolution of the super-resolved image reaches $320\times320$. We train the framework GIRAFFE for more than $1.3$ million iterations.
%
\subsection{Additional experimental results} \label{sec:more_exp}
In this section, we present the results on more complicated real-world scenes in \ref{sec:real_data} and investigate the robustness of the framework to the underlying camera pose distribution for the multi-view observations used for training in \ref{sec:camera_noise}.

\subsubsection{Results on real-world outdoor scenes}\label{sec:real_data}
In this section, we provide the results for \name on two more complicated real scenes, which are collected from \href{https://earth.google.com}{Google Earth}\footnote{https://earth.google.com}, namely \emph{Devil Tower} and \emph{Bagana Volcano}, respectively. 
After training on the collected data, \name achieves quantitative results shown in Table \ref{tab:real_world}. 

\begin{table*}[htbp]
\scriptsize
\centering
\caption{{Numerical results on real-world data.}}
\begin{tabular}{lccc}
\toprule
Variant     & Data            & SIFID-MV $\downarrow$ & Diversity-MV $\uparrow$ \\ 
\midrule
SinGRAV     & Devil Tower                 & 0.1940    & 0.5801 \\      
SinGRAV     & Bagana Volcano               & 0.1480    & 0.7359 \\ 
SinGRAV (wo. depth sup.) & Bagana Volcano     & 0.1710    & 0.8391 \\ \bottomrule
\end{tabular}
\label{tab:real_world}%
\end{table*}
\begin{figure*}
    \includegraphics[width=\linewidth]{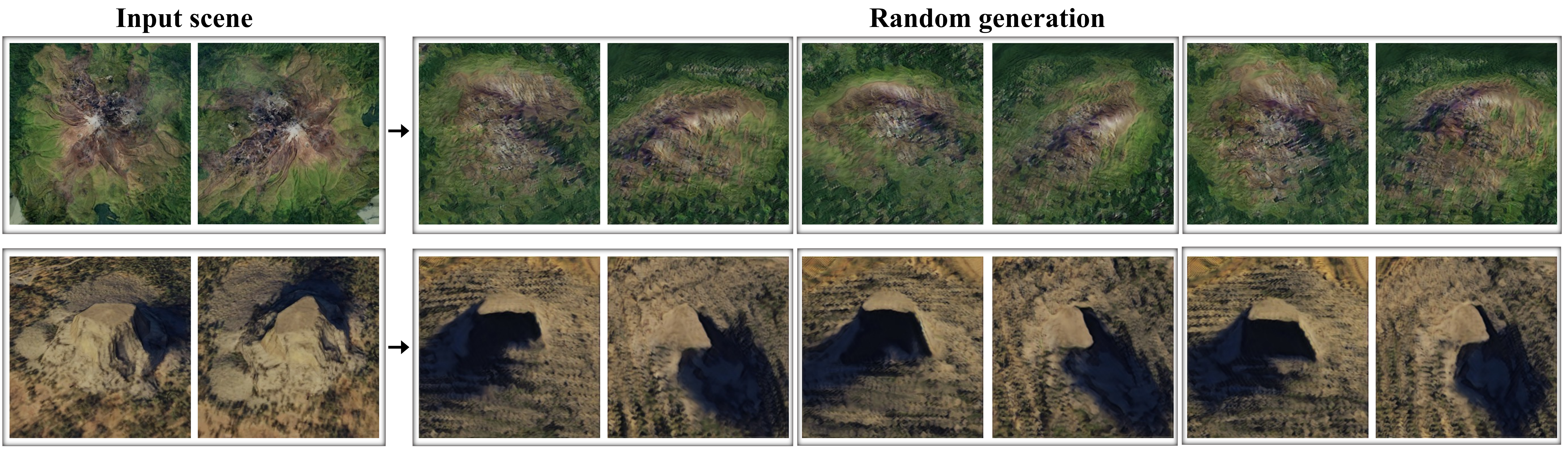}
    \caption{
    \revise{Random scene samples from \name trained on real-world data.
    At each row, we show two views of the input scene on the left, followed by three diverse scenes generated by \name on the right.
    Each scene is rendered with the same two cameras.}
    }
    \label{fig:real_world}
\end{figure*}

The first row and the second row in Table \ref{tab:real_world} show that the numerical results on real-world scenes approaches those demonstrated in Table \ref{tab:ablation}. 
This suggests that \name also produces reasonable results on more complex real-world scenes. 
On the qualitative perspective, we notice that, while our method produces plausible generation visual results, the fine textures of the real-world scene indeed challenge the proposed voxel-based framework as also discussed in the main paper.
Note these real-world results are obtained using the default hyper-parameter setting, without ad hoc fine-tuning.

\begin{figure*}[h]
\centering
    \includegraphics[width=\linewidth]{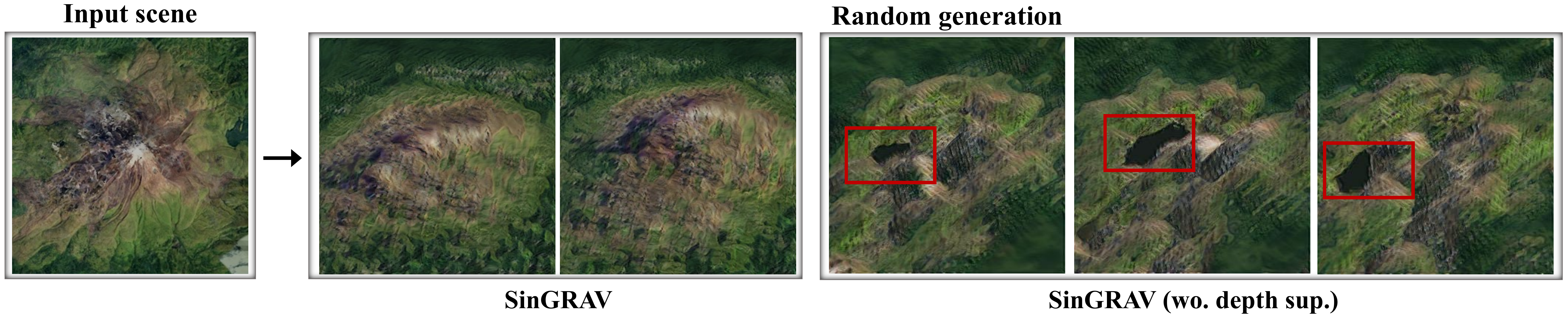}
    \caption{
    \revise{Influence of the depth supervisions on the real-world scene. 
    Two generated scenes are provided for \name and three results are provided for \name (wo. depth sup.). 
    Red boxes highlight the implausible geometry \--- holes in the results of \name (wo. depth sup.), that are caused by the lack of depth supervision.}
    }
    \label{fig:ablation:real_data}
\end{figure*}

In addition, 
we also investigate the influence of the depth supervision on one of the above scenes \--- \emph{Bagana Volcano}, and report the numerical results in the third row of Table \ref{tab:real_world}. 
It can be seen that the numerical results drop slightly, which is similar to the trend shown in Table \ref{tab:ablation}. 
Again, despite the slight drop in numerical results, 
totally eliminating depth supervisions on this real-world data also drastically affects the plausibility of the spatial arrangements in the results, 
which can be observed from the qualitative results shown in Figure \ref{fig:ablation:real_data}.
Nevertheless, the depth data obtained with COLMAP in our default setting is sufficient for producing reasonable geometric structures in the results, as also stated in the main paper.

 \label{sec:camera_noise}
\begin{figure*}
\centering
    \includegraphics[width=0.7\linewidth]{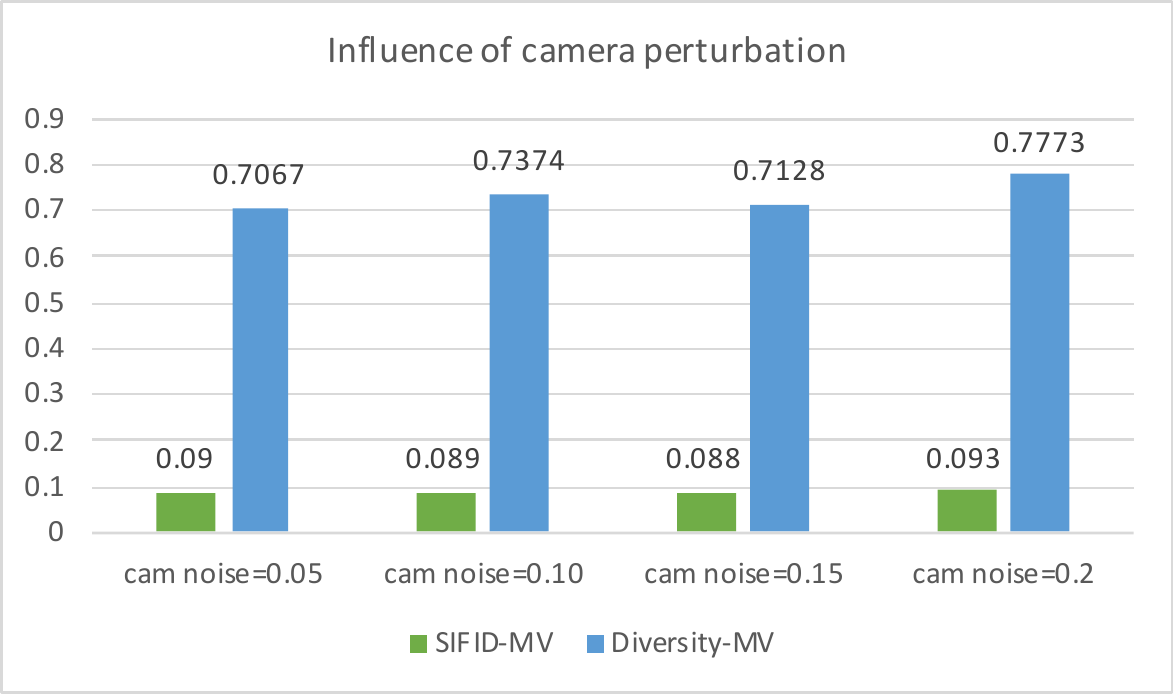}
    \caption{
    Performance for \name trained on data obtained with perturbed camera poses. Within a reasonable perturbation range ($\pm20$ cm), the performance is stable, denoting that the \name is robust to the reasonable perturbations for camera pose distribution during the data acquisition. 
    }
    \label{fig:ablation_cam_noise}
\end{figure*}

\subsubsection{Robustness to camera distribution}  
In real-world scenarios,
it is likely that the cameras would deviate from the planned trajectories to some extent, i.e., the hemisphere camera distribution may not be guaranteed under a less controlled setting.
Nevertheless,
we shall show that \name does NOT necessitate an absolutely strict distribution for the training cameras, and can work properly if two criteria can be met: 
a) \emph{high observation completeness}, which suggests the scene of interest needs to be covered as much as possible,
and is a reasonable assumption with the prevalent "render-and-discriminate" framework; 
b) \emph{consistent multi-scale internal distribution}, 
which means the captured images need to share a similar internal distribution at each scale for learning the generative model.

To account for such factors, 
we investigate the robustness of \name under such less controlled environments by simulating with perturbed camera poses.
Specifically, 
we randomly add noise drawn from a Gaussian distribution with different standard deviations to the camera poses during the multi-view images acquisition.
We conduct the study on \emph{Stonehenge}, 
of which the results are demonstrated in Figure \ref{fig:ablation_cam_noise}.
Figure \ref{fig:ablation_cam_noise} shows that the scores for SIFID-MV and Diversity-MV stay similar when the scale of the Gaussian noise changes within a range of $\left[-20, +20\right]$ cm.
Experimental results show that \name is robust to adequate perturbations of the camera poses, which means that \name just needs the camera poses to come from a roughly satisfying distribution rather than strictly demands that the camera poses come from exactly controlled settings.

\subsection{Video demonstrations} \label{sec:video_results}
\revise{Here we give outlines for video results we provide in the supplementary videos. 
In Supplementary Video, we organize the results in the following manner:}
\begin{itemize}
    \item Qualitative results on various datasets from \name.
    \item Qualitative results from \name-derived variants.
    \item Qualitative results from baselines including GRAF \cite{graf}, pi-GAN \cite{pi-gan} and GIRAFFE \cite{giraffe}.
    \item Qualitative results for applications.
\end{itemize}


\end{document}